\pdfoutput=1

\documentclass[11pt]{article}

\usepackage{emnlp2021}

\usepackage{times}
\usepackage{latexsym}

\usepackage[T1]{fontenc}

\usepackage[utf8]{inputenc}

\usepackage{microtype}

\usepackage{amsmath}
\usepackage{hyperref}
\usepackage{tikz}
\usepackage{pgfplots}
\pgfplotsset{compat=1.16}
\usepackage{multirow}
\usepackage{multicol}
\usepackage{caption}
\usepackage{graphicx}
\usepackage{tabu}
\usepackage{makecell}
\usepackage{soul}
\usepackage{booktabs}
\usepackage{array}
\usepackage{ragged2e}
\usepackage{subcaption}
\usepackage{footmisc}
\usepackage{algorithm}
\usepackage[noend]{algpseudocode}
\usepackage{xpatch}
\makeatletter
\xpatchcmd{\algorithmic}{\itemsep\z@}{\itemsep=-0.1ex plus2pt}{}{}
\makeatother
\renewcommand{\algorithmicrequire}{\textbf{Input:}}
\renewcommand{\algorithmicensure}{\textbf{Output:}}
\title{A Strong Baseline for Query Efficient Attacks in a Black Box Setting}

\author{Rishabh Maheshwary\thanks{\; Equal Contribution} , Saket Maheshwary\footnotemark[1] \and Vikram Pudi\\
Data Sciences and Analytics Center, Kohli Center on Intelligent Systems \\
International Institute of Information Technology, Hyderabad, India \\ {\{\texttt{rishabh.maheshwary@research.iiit.ac.in, vikram@iiit.ac.in}\}}}

\begin{document}
\maketitle
\begin{abstract}
Existing black box search methods have achieved high success rate in generating adversarial attacks against NLP models.
However, such search methods are inefficient as they do not consider the amount of queries required to generate adversarial attacks. Also, prior attacks do not maintain a consistent search space while comparing different search methods. In this paper, we propose a query efficient attack strategy to generate plausible adversarial examples on text classification and entailment tasks. Our attack jointly leverages attention mechanism and locality sensitive hashing (LSH) to reduce the query count. We demonstrate the efficacy of our approach by comparing our attack with \emph{four} baselines across \emph{three} different search spaces. Further, we benchmark our results across the same search space used in prior attacks. In comparison to attacks proposed, on an average, we are able to reduce the query count by $75\%$ across all datasets and target models. We also demonstrate that our attack achieves a higher success rate when compared to prior attacks in a limited query setting.
\end{abstract}

\section{Introduction}
In recent years, Deep Neural Networks (DNNs) have achieved high performance on a variety of tasks~\cite{yang2016hierarchical,goodfellow2016deep,kaul2017autolearn,maheshwary2017deep,maheshwary2018matching,devlin2018bert}. However,
 prior studies~\citep{szegedy2013intriguing,papernot2017practical} have shown evidence that DNNs are vulnerable to \emph{adversarial examples} --- inputs generated by slightly changing the original input. Such changes are imperceptible to humans but they deceive DNNs, thus raising serious concerns about their utility in real world applications. Existing NLP attack methods are broadly classified into \emph{white box attacks} and \emph{black box attacks}. White box attacks require access to the target model's parameters, loss function and gradients to craft an adversarial attack. Such attacks are computationally expensive and require knowledge about the internal details of the target model which are not available in most real world applications. Black box attacks crafts adversarial inputs using only the confidence scores or class probabilities predicted by the target model.
Almost all the prior black box attacks consists of two major components $(1)$ \emph{search space} and $(2)$ \emph{search method}.

A \emph{search space} is collectively defined by a set of transformations (usually synonyms) for each input word and a set of constraints (e.g., minimum semantic similarity, part-of-speech (POS) consistency). The synonym set for each input word is generated either from the nearest neighbor in the counter-fitted embedding space or from a lexical database such as HowNet~\cite{dong2003hownet} or WordNet~\cite{miller1995wordnet}. The search space is variable and can be altered by either changing the source used to generate synonyms or by relaxing any of the above defined constraints.

A \emph{search method} is a searching algorithm used to find adversarial examples in the above defined search space. Given an input with $\mathcal{W}$ words and each word having $\mathcal{T}$ possible substitutes, the total number of perturbed text inputs is $(\mathcal{T}+1)^\mathcal{W}$. Given this exponential size, the search algorithm must be efficient and exhaustive enough to find optimal adversarial examples from the whole search space.

Black box attacks proposed in~\cite{alzantot2018generating,zang2020word} employ combinatorial optimization procedure as a search method to find adversarial examples in the above defined search space. Such methods are extremely slow and require massive amount of queries to generate adversarial examples. Attacks proposed in~\cite{ren2019generating,jin2019bert} search for adversarial examples using word importance ranking which first rank the input words and than substitutes them with similar words. The word importance ranking scores each word by observing the change in the confidence score of the target model after that word is removed from the input (or replaced with <UNK> token). Although, compared to the optimization based methods, the word importance ranking methods are faster, but it has some major drawbacks -- $(1)$ each word is ranked by removing it from the input (or replacing it with a <UNK> token) which therefore alter the semantics of the input during ranking, $(2)$ it not clear whether the change in the confidence score of the target model is caused by the removal of the word or the modified input and $(3)$ this ranking mechanism is inefficient on input of larger lengths. 

In general, their exists a trade off between the attack success rate and the number of queries. A high query search method generates adversarial examples with high success rate and vice-versa. All such prior attacks are not efficient and do not take into consideration the number of queries made to the target model to generate an attack. Such attacks will fail in real world applications where there is a constraint on the number of queries that can be made to the target model.

To compare a new search method with previous methods, the new search method must be benchmarked on the same search space used in the previous search methods. However, a study conducted in~\cite{yoo2020searching} have shown that prior attacks often modify the search space while evaluating their search method. This does not ensure a fair comparison between the search methods because it is hard to distinguish whether the increase in attack success rate is due to the improved search method or modified search space. For example, ~\cite{jin2019bert} compares their search method with~\citep{alzantot2018generating} where the former uses Universal Sentence Encoder (USE)~\cite{cer2018universal} and the latter use language model as a constraint. Also, all the past works evaluate their search methods only on a single search space. 
In this paper\footnote{Code and adversarial examples available at: \href{https://github.com/RishabhMaheshwary/query-attack}{https://github.com/RishabhMaheshwary/query-attack}}, we address the above discussed drawbacks through following contributions:
\begin{enumerate}
    \item We introduce a \emph{novel ranking mechanism} that takes significantly less number of queries by jointly leveraging word attention scores and LSH to rank the input words; without altering the semantics of the input.
    \item We call for unifying the evaluation setting by benchmarking our search method on the same search space used in the respective baselines. Further, we evaluate the effectiveness of our method by comparing it with \emph{four} baselines across \emph{three} different search spaces.
    \item On an average, our method is $50\%$ faster as it takes $75\%$ lesser queries than the prior attacks while compromising the attack success rate by less than $2\%$. Further, we demonstrate that our search method has a much higher success rate than compared to baselines in a limited query setting.
\end{enumerate}
\section{Related Work}
\begin{figure*}
    \centering
    \includegraphics[width=16.0cm,height=6.0cm]{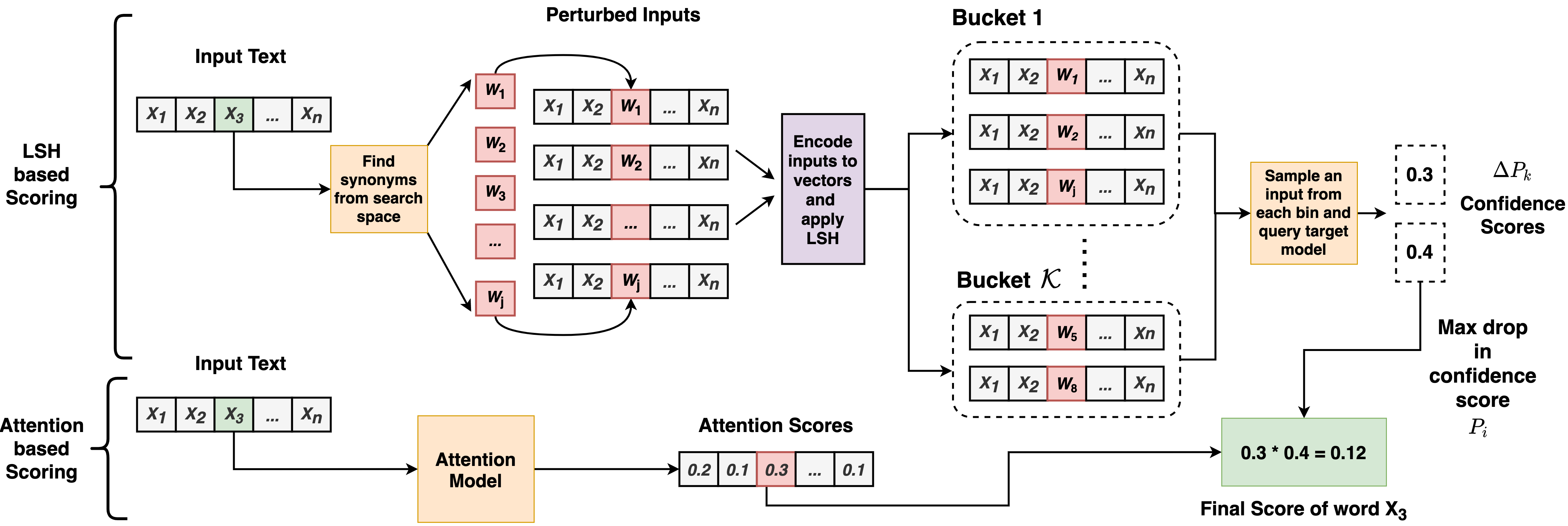}
    \caption{Scoring of each input word using attention mechanism and Locality Sensitive Hashing (LSH).}
    \label{fig:arch}
\end{figure*}

\subsection{White Box Attacks} 
This category requires access to the gradient information to generate adversarial attacks. Hotflip~\cite{ebrahimi2017hotflip} flips characters in the input using the gradients of the one hot input vectors. \citeauthor{liang2017deep} used gradients to perform insertion, deletion and replacement at character level. Later~\cite{li2018textbugger} used the gradient of the loss function with respect to each word to find important words and replaced those with similar words. Following this,~\cite{wallace2019universal} added triggers at the start of the input to generate adversarial examples. Although these attacks have a high success rate, but they require knowledge about the model parameters and loss function which is not accessible in real word scenarios.
\subsection{Black Box Attacks} 
Existing black box attacks can be classified into combinatorial optimization based attacks and greedy attacks. Attack proposed in~\cite{alzantot2018generating} uses Genetic algorithm as a search method to search for optimal adversarial examples in the search space. On similar lines,~\cite{zang2020word} uses Particle Swarm Optimization procedure to search for adversarial examples. Recently,~\cite{maheshwary2021generating} crafted adversarial attacks using Genetic algorithm in a hard label black box setting. Although, such methods have a high success rate but they are extremely slow and takes large amount of queries to search for optimal adversarial examples.

Greedy black box attacks generates adversarial attacks by first finding important words, which highly impacts the confidence score of the target model and than replacing those words with similar words. Search method proposed in~\cite{ren2019generating} used word saliency and the target model confidence scores to rank words and replaced them with synonyms from WordNet~\cite{miller1995wordnet}. Although such a ranking mechanism is exhaustive, but it is not query efficient to rank the words. Inspired by this,~\citep{jin2019bert} proposed TextFooler which ranks word only based upon the target model's confidence and replaces them with synonyms from the counter-fitted embedding space~\cite{mrkvsic2016counter}. This ranking mechanism have a low attack success rate and is not exhaustive enough to search for adversarial examples with low perturbation rate.

Some prior works~\cite{garg2020bae,li2020bert,maheshwary2020context,li2020contextualized} have used masked language models to generate word replacements instead of using synonyms. However, all these methods follow a ranking mechanism similar to TextFooler. Moreover, as shown in~\cite{yoo2020searching} most of the black box methods described above do not maintain a consistent search space while comparing their method with other search methods.

\subsection{Locality Sensitive Hashing} 
LSH has been used in various NLP applications in the past. ~\citeauthor{ravichandran2005randomized} used LSH for clustering nouns, ~\cite{van2010online} extended it for streaming data. Recently,~\cite{kitaev2020reformer} used LSH to reduce the computation time of self attention mechanism. There have been many variants of LSH, but in this paper we leverage the LSH method proposed in~\cite{charikar2002similarity}. 

\section{Proposed Approach}
Given a target model $\bf{F}: \mathcal{X}\rightarrow \mathcal{Y}$, that maps the input text sequence $\mathcal{X}$ to a set of class labels $\mathcal{Y}$. Our goal is to generate an adversarial text sequence $\mathcal{X_{ADV}}$ that belongs to any class in $\mathcal{Y}$ except the original class of $\mathcal{X}$ i.e. $\bf{F}(\mathcal{X}) \ne\bf{F}(\mathcal{X_{ADV}})$. The input $\mathcal{X_{ADV}}$ must be generated by substituting the input words with their respective synonyms from a chosen search space.\\
Our search method consists of two steps $(1)$ Word Ranking -- ranks all words in the input text and $(2)$ Word Substitution -- substitutes input words with their synonyms in the order-of-rank (step $1$).
\subsection{Word Ranking}
Recent studies~\citep{niven2019probing} have shown evidence that certain words in the input and their replacement can highly influence the final prediction of DNNs. Therefore, we score each word based upon, $(1)$ \emph{how important it is for the final prediction} and  $(2)$ \emph{how much its replacement with a similar word can alter the final prediction} of the target model.
We use \emph{attention mechanism} to select important words for classification and employ \emph{LSH} to capture the impact of replacement of each word on the prediction of target model. Figure \ref{fig:arch} demonstrates the working of the word ranking step.
\subsubsection{Attention based scoring}
Given an input $\mathcal{X}$, this step assigns high score to those influential words which impact the final outcome. The input sequence $\mathcal{X} = \{x_1,x_2..x_n\}$ is passed through a pre-trained attention model $F_{attn}$ to get attention scores $\alpha_i$ for each word $x_i$. The scores are computed using Hierarchical Attention Networks (HAN)~\citep{yang2016hierarchical} and Decompose Attention Model (DA)~\citep{parikh2016decomposable} for text classification and entailment tasks respectively. Note, instead of querying the target model every time to score a word, this step scores all words together in a single pass (inferring the input sample by passing it through attention model), thus, significantly reducing the query count. Unlike prior methods, we do not rank each word by removing it from the input (or replacing it with a UNK token), preventing us from altering semantics of the input.
\subsubsection{LSH based Scoring}
This step assigns high scores to those words whose replacement with a synonym will highly influence the final outcome of the model. It scores each word based on the change in confidence score of the target model, when it is replaced by its substitute (or synonym) word. But computing the change in confidence score for each synonym for every input word, significantly large number of queries are required. Therefore, we employ LSH to solve this problem. LSH is a technique used for finding nearest neighbours in high dimensional spaces. It takes an input, a vector $x$ and computes its hash $h({x})$ such that similar vectors gets the same hash with high probability and dissimilar ones do not. LSH differs from cryptographic hash methods as it aims to maximize the collisions of similar items. We leverage Random Projection Method (RPM)~\cite{charikar2002similarity} to compute the hash of each input text.
\subsubsection{Random Projection Method (RPM)}
Let us assume we have a collection of vectors in an $m$ dimensional space $R^m$. Select a family of hash functions by randomly generating a spherically symmetrical random vector $r$ of unit length from the $m$ dimensional space. Then the hash function $h_r(u)$ is defined as:

\begin{equation}
\label{eq:1}
     \text{$h_r(u)$} =  \begin{cases} 
      0 & r.u < 0 \\
      1 & r.u \geq 0 
   \end{cases}
\end{equation}
Repeat the above process by generating $d$ random unit vectors $\{r_0,r_1..r_d\}$ in $R^m$. The final hash $\Bar{u}$ for each vector $u$  is determined by concatenating the result obtained using \label{eq:1} on all $d$ vectors. 
\begin{equation}
    \label{eq:2}
    \Bar{u} = \{h_{r1}(u),h_{r2}(u)..h_{rd}(u)\}.
\end{equation}
The hash of each vector $u$ is represented by a sequence of bits and two vectors having same hash are mapped to same bucket in the hash table. Such a process is very efficient in finding nearest neighbor in high dimensional spaces as the hash code is generated using only the dot product between two matrices. Also, it is much easier to implement and simple to understand when compared to other nearest neighbour methods.
We use the above process to score each word as follows:
\begin{enumerate}
    \item First, an input word $x_i$ is replaced with every synonym from its synonym set $S(x_i)$ resulting in perturbed sequence $\mathcal{X}_{ij} = \{x_1...w_j...x_n\}$, where $i$ is the substituted index and $j$ is the $jth$ synonym from $S(x_i)$ i.e. $w_j \in S(x_i)$. Perturbed sequences not satisfying the search space constraints (Table \ref{table1}) are filtered.
    \item The remaining perturbed inputs are passed through a sentence encoder (USE) which returns a vector representation $\mathcal{V}_j$ for each perturbed input. Then we use LSH as described above to compute the hash of each vector. The perturbed inputs having the same hash are mapped to same bucket of the hash table.
    \begin{gather}
            \mathcal{V}_j = \mathbf{encode}(\mathcal{X}_{ij}) \quad \forall j; \ j \in [1,\mathcal{T}]\\
        \mathcal{B} = \mathbf{LSH}(\mathcal{V}_j,\mathcal{X}_{ij}) \quad \forall j; \ j \in [1,\mathcal{T}]
    \end{gather}
were $\mathcal{T}$ are the number of perturbed inputs obtained for each word and $\mathcal{B}  = \{b_0,b_1...b_\mathcal{K}\}$ are the buckets obtained after LSH in a hash table, $\mathcal{K}$ being the number of buckets.
    \item As each bucket contains similar perturbed inputs, a perturbed input is sampled randomly from each bucket and is passed to the target model $\mathbf{F}$. The maximum change in the confidence score of the target model among all these fed inputs is the score for that index.
    \begin{gather}
        \mathcal{V}_k^*, \mathcal{X}_k^* = sample(b_k) \ \forall k; \ k \in [0,\mathcal{K}]\\
        \Delta P_k = \mathbf{F}(\mathcal{X}) - \mathbf{F}(\mathcal{X}_k^*) \ \forall k; \ k \in [0,\mathcal{K}]\\
        P_i = \max (\Delta P_k) \quad \quad k \in [0,\mathcal{K}]
    \end{gather}
\end{enumerate}
The steps $1$ to $3$ are repeated for all the indices in $\mathcal{X}$. LSH maps highly similar perturbed inputs to the same bucket, and as all such inputs are similar, they will impact the target model almost equally. Therefore, instead of querying the target model $\textbf{F}$ for every input in the bucket, we sample an input randomly from each bucket and observe its impact on the target model $\textbf{F}$. This will reduce the query count from being proportional to number of synonyms of each word to minimum number of buckets obtained after LSH. 
\subsubsection{False Negative error rate of LSH}
Although LSH is efficient for finding nearest neighbour, there is still a small probability that similar perturbed inputs get mapped to different buckets. Therefore to reduce this probability, we conduct multiple rounds of hashing, $L = 15$, each round with a different family of hash functions and choose the round which has the most collisions i.e. round having minimum buckets.~\citep{charikar2002similarity} establishes an upper bound on the false negative error rate of LSH i.e two highly similar vectors are mapped to different buckets.
The upper bound on the false negative error rate of LSH given by (for more details refer~\cite{charikar2002similarity})
\begin{gather}
\label{eq:lsh}
     (1-P^d)^L \\
    where \  P \ = \  1 - \dfrac{\theta(u,v)}\pi
\end{gather}
This shows that for given values of $L$ and $d$, LSH maps similar vectors to the same bucket with very high probability.
As LSH maps similar inputs to the same bucket with high probability, it cuts down the synonym search space drastically, thus reducing the number of queries required to attack. The dimension of the hash function $d$ used in equation \ref{eq:2} is set to $5$ and is same across all rounds.

\subsubsection{Final Score Calculation}
After obtaining the attention scores $\alpha_i$ and the scores from synonym words $P_i$ for each index (calculated using LSH), we multiply the two to get the final score $score_{i} = \alpha_i * P_i$ for each word. All the words are sorted in descending order based upon the score $score_i$. The algorithm of the word ranking step is provided in the appendix \ref{sec:appendix}.
\subsection{Word Substitution}
We generate the final adversarial example for the input text by perturbing the words in the order retrieved by the word ranking step.
For each word $x_{i}$ in $\mathcal{W}$, we follow the following steps.
\begin{enumerate}
  \item  Each synonym $w_j$ from the synonym set $S(x_i)$ is substituted for $x_i$ which results in a perturbed text $X'_{j} = \{x_1..w_j..x_n\}$. The perturbed texts which do not satisfy the constraints imposed on the search space (Table \ref{table1}) are filtered (Algorithm $1$, lines $1-7$).
  
  \item The remaining perturbed sequence(s) are fed to the target model $\bf{F}$ to get the class label $y_{new}$ and its corresponding probability score $P_{new}$. The perturbed sequence(s) which alters the  original class label $y_{orig}$ is chosen as the final adversarial example $\mathcal{X_{ADV}}$. In case the original label does not change, the perturbed sequence which has the minimum $P_{new}$ is chosen (Algorithm $1$, lines $7-14$). 
 \end{enumerate}
The steps $1-2$ are repeated on the chosen perturbed sequence for the next ranked word. Note, we only use LSH in the word ranking step, because in ranking we need to calculate scores for all the words in the input and so we need to iterate over all possible synonyms of every input word. However, in the word substitution step we replace only one word at a time and the substitution step stops when we get an adversarial example. As the number of substitutions are very less (see perturbation rate in Table \ref{table6}) to generate adversarial example, the substitution step iterate over very less words when compared to ranking step. 
\begin{algorithm}[h!]
\caption{Word Substitution}
\algorithmicrequire{ Test sample $\mathcal{X}$, Ranked words $\mathcal{W}$ }\\
\algorithmicensure{ Adversarial text $\mathcal{X}_{ADV}$}
\begin{algorithmic}[1]
\State $\mathcal{X}_{ADV} \gets \mathcal{X}$
\State $y_{orig}\ ,\ P_{orig} \gets \bf{F}(\mathcal{X_{ADV}})$
\State $P_{best} \gets P_{orig}$
\For {$(score_i,x_{i})\ in\ \mathcal{W}$}
\State ${S} \gets\ Synonyms(x_{i})$
\For {$w_{j}\ in\ {S}$}
 \State $X'_{j} \gets \ Replace\ x_i\ with\ w_j $
\State  $y_{new}\ , \ P_{new} \gets \textbf{F}({X'_j})$
\If{$y_{new} \ne y_{orig}$}
\State $\mathcal{X_{ADV}} \gets X'_j$\
\State \Return $\mathcal{X_{ADV}}$
\EndIf
\If{$P_{new} < P_{best}$}
\State $P_{best} \gets P_{new}$
\State $\mathcal{X_{ADV}} \gets X'_j$
\EndIf
\EndFor
\EndFor
\State \Return $\mathcal{X}_{ADV}$
\end{algorithmic}
\end{algorithm}
\section{Experiments}
\subsection{Datasets and Target Models}
We use \emph{IMDB} --- A document level sentiment classification dataset for movie reviews~\cite{maas2011learning} and \emph{Yelp Reviews} --- A restaurant review dataset~\citep{zhang2015character}, for classification task. We use \emph{MultiNLI} --- A natural language inference dataset~\citep{williams2017broad} for entailment task. We attacked \emph{WordLSTM}~\citep{hochreiter1997long} and \emph{BERT-base-uncased}~\citep{devlin2018bert} for evaluating our attack strategy on text classification and entailment tasks. For WordLSTM, we used a single layer bi-directional LSTM with $150$ hidden units, a dropout of $0.3$ and $200$ dimensional GloVe~\cite{pennington2014glove} vectors. Additional details are provided in appendix \ref{sec:appendix}.
\subsection{Search Spaces and Baselines}
We compare our search method with four baselines across three different search spaces. Also, while comparing our results with each baselines we use the same search space as used in that baseline paper. The details of search spaces are shown in Table \ref{table1}.\\
\emph{PSO:}~\citep{zang2020word} It uses particle swarm optimization algorithm as a search method and uses HowNet~\cite{dong2006hownet} as search space.\\
\emph{TextFooler:}~\citep{jin2019bert} It finds important words and replace them with synonyms from counter-fitted embeddings~\cite{mrkvsic2016counter}.\\
\emph{PWWS:}~\citep{ren2019generating} It takes word saliency score into account to rank words and uses WordNet~\citep{miller1995wordnet} for synonym substitution.\\
\emph{Genetic Attack:} It crafts examples using a population based algorithm~\citep{alzantot2018generating}.
\begin{table}[h!]
\centering
\resizebox{0.49\textwidth}{!}{%
{\renewcommand{\arraystretch}{1.2}
\begin{tabular}{|c|c|c|}
\hline
\textbf{Baseline} & \textbf{Transformation} & \textbf{Constraint} \\ \hline
Genetic Attack & {Counter-fitted} & Word similarity, LM score\\ \hline
TextFooler & embeddings &  USE = $0.84$, POS consistency \\ \hline
PSO & HowNet & USE = $0.84$, POS consistency \\ \hline
PWWS & WordNet & POS consistency \\ \hline
\end{tabular}}}
  \caption{Baselines and their search spaces}
  \label{table1}
\end{table}
\begin{table*}[h!]

\centering
\begin{subtable}[h]{0.49\textwidth}
\centering
\resizebox{1.0\textwidth}{!}{%
{\renewcommand{\arraystretch}{1.2}
\begin{tabular}{|c|c|c|c|c|c|c|c|}
\hline
\multirow{2}{*}{\textbf{Model}} & \multirow{2}{*}{\textbf{Attack}} & \multicolumn{2}{c|}{\textbf{IMDB}} & \multicolumn{2}{c|}{\textbf{Yelp}} & \multicolumn{2}{c|}{\textbf{MNLI}} \\ \cline{3-8} 
 &  & \textbf{Qrs} & \textbf{Suc\%} & \textbf{Qrs} & \textbf{Suc\%} & \textbf{Qrs} & \textbf{Suc\%} \\ \hline
\multirow{2}{*}{\textbf{BERT}} & PSO &  81350.6 & \bf{99.0} & 73306.6 & \bf{93.2} & 4678.5 & \bf{57.97}  \\
 & Ours & \bf{737} & 97.4  & \bf{554.2} & 91.6 & \bf{97.2} & 56.1 \\ \hline
\multirow{2}{*}{\textbf{LSTM}} & PSO  & 52008.7 & \bf{99.5}  & 43671.7 & \bf{95.4}  & 2763.3 & \bf{67.8} \\
 & Ours  & \bf{438.1} & \bf{99.5} & \bf{357.6} & 94.75  & \bf{79.8} & 66.4 \\ \hline
\end{tabular}}}
\caption{Comparison with PSO.}
\label{table2a}
\end{subtable}
\hfill
\begin{subtable}[h]{0.49\textwidth}
\centering
\resizebox{1.0\textwidth}{!}{%
{\renewcommand{\arraystretch}{1.2}
\begin{tabular}{|c|c|c|c|c|c|c|c|}
\hline
\multirow{2}{*}{\textbf{Model}} & \multirow{2}{*}{\textbf{Attack}} & \multicolumn{2}{c|}{\textbf{IMDB}} & \multicolumn{2}{c|}{\textbf{Yelp}} & \multicolumn{2}{c|}{\textbf{MNLI}} \\ \cline{3-8} 
 & & \textbf{Qrs} & \textbf{Suc\%} & \textbf{Qrs} & \textbf{Suc\%} & \textbf{Qrs} & \textbf{Suc\%}  \\ \hline
\multirow{2}{*}{\textbf{BERT}} & Gen & 7944.8 & 66.3 & 6078.1 & \bf{85.0} & 1546.8 & \bf{83.8} \\
 & Ours  & \bf{378.6} & \bf{71.1} & \bf{273.7} & {84.4}  & \bf{43.4} & 81.9 \\ \hline
\multirow{2}{*}{\textbf{LSTM}} & Gen & 3606.9 & 97.2 & 5003.4 & \bf{96.0} & 894.5 & \bf{87.8} \\
 & Ours & \bf{224} & \bf{98.5} & \bf{140.7} & 95.4 & \bf{39.9} & 86.4  \\ \hline
\end{tabular}}}
\caption{Comparison with Genetic Attack.}
\label{table2b}
\end{subtable}
\begin{subtable}[h]{0.49\textwidth}
\centering
\resizebox{1.0\textwidth}{!}{%
{\renewcommand{\arraystretch}{1.2}
\begin{tabular}{|c|c|c|c|c|c|c|c|}
\hline
\multirow{2}{*}{\textbf{Model}} & \multirow{2}{*}{\textbf{Attack}} & \multicolumn{2}{c|}{\textbf{IMDB}} & \multicolumn{2}{c|}{\textbf{Yelp}} & \multicolumn{2}{c|}{\textbf{MNLI}} \\ \cline{3-8} 
 &  & \textbf{Qrs} & \textbf{Suc\%} & \textbf{Qrs} & \textbf{Suc\%}  & \textbf{Qrs} & \textbf{Suc\%} \\ \hline
\multirow{2}{*}{\textbf{BERT}} & PWWS & 1583.9 & \bf{97.5} & 1013.7 & \bf{93.8} & 190 & \bf{96.8}\\
 & Ours & \bf{562.9} & 96.4  & \bf{366.2} & 92.6 & \bf{66.1} & 95.1 \\ \hline
\multirow{2}{*}{\textbf{LSTM}} & PWWS  & 1429.2 & \bf{100.0} & 900.0 & \bf{99.1}  & 160.2 & \bf{98.8}\\
 & Ours & \bf{473.8} & \bf{100.0} & \bf{236.3} & \bf{99.1} & \bf{60.1}  & 98.1 \\ \hline
\end{tabular}}}
\caption{Comparison with PWWS.}
\label{table2c}
\end{subtable}
\hfill
\begin{subtable}[h]{0.49\textwidth}
\centering
\resizebox{1.0\textwidth}{!}{%
{\renewcommand{\arraystretch}{1.2}
\begin{tabular}{|c|c|c|c|c|c|c|c|}
\hline
\multirow{2}{*}{\textbf{Model}} & \multirow{2}{*}{\textbf{Attack}} & \multicolumn{2}{c|}{\textbf{IMDB}} & \multicolumn{2}{c|}{\textbf{Yelp}} & \multicolumn{2}{c|}{\textbf{MNLI}} \\ \cline{3-8} 
 & & \textbf{Qrs}  & \textbf{Suc\%} & \textbf{Qrs} & \textbf{Suc\%}  & \textbf{Qrs} & \textbf{Suc\%} \\ \hline
\multirow{2}{*}{\textbf{BERT}} & TF  & 1130.4 & \bf{98.8} & 809.9 & \bf{94.6} & 113 & 85.9 \\
 & Ours & \bf{750} & 98.4 & \bf{545.5} & 93.2 & \bf{100} & \bf{86.2}\\ \hline
\multirow{2}{*}{\textbf{LSTM}} & TF  & 544 & \bf{100.0}  & 449.4 & \bf{100.0}  & 105 & 95.9 \\
 & Ours  & \bf{330} & \bf{100.0}  & \bf{323.7} & \bf{100.0}  & \bf{88} & \bf{96.2} \\ \hline
\end{tabular}}}
\caption{Comparison with TextFooler (TF).}
\label{table2d}
\end{subtable}
\caption{Result comparison. Succ\% is the attack success rate and Qrs is the average query count. Note as each baseline uses a different search space, our method will yield different results when comparing with each baseline.}
\label{table2}
\end{table*}

\begin{table*}[h!]

\centering
\begin{subtable}[h]{0.49\textwidth}

\centering
\resizebox{1.0\textwidth}{!}{%
{\renewcommand{\arraystretch}{1.2}
\begin{tabular}{|c|c|c|c|c|c|c|c|}
\hline
\multirow{2}{*}{\textbf{Model}} & \multirow{2}{*}{\textbf{Attack}} & \multicolumn{2}{c|}{\textbf{IMDB}} & \multicolumn{2}{c|}{\textbf{Yelp}} & \multicolumn{2}{c|}{\textbf{MNLI}} \\ \cline{3-8} 
 &  & \textbf{Pert\%} & \textbf{I\%} & \textbf{Pert\%} & \textbf{I\%} & \textbf{Pert\%} & \textbf{I\%} \\ \hline
\multirow{2}{*}{\textbf{BERT}} & PSO & 4.5 & 0.20 & 10.8 & 0.30 & 8.0 & 3.5 \\
 & Ours & \bf{4.2} & \bf{0.10} & \bf{7.8} & \bf{0.15} & \bf{7.1} & \bf{3.3} \\ \hline
\multirow{2}{*}{\textbf{LSTM}} & PSO & 2.2 & 0.15 & 7.7 & 0.27  &  \bf{6.7} & \bf{1.27} \\
 & Ours & \bf{2.0} & \bf{0.11} & \bf{4.9} & \bf{0.15} & 6.8 & 1.3 \\ \hline
\end{tabular}}}
  \caption{Comparison with PSO.}
  \label{table3a}
\end{subtable}
\hfill
\begin{subtable}[h]{0.49\textwidth}

\centering
\resizebox{1.0\textwidth}{!}{%
{\renewcommand{\arraystretch}{1.2}
\begin{tabular}{|c|c|c|c|c|c|c|c|}
\hline
\multirow{2}{*}{\textbf{Model}} & \multirow{2}{*}{\textbf{Attack}} & \multicolumn{2}{c|}{\textbf{IMDB}} & \multicolumn{2}{c|}{\textbf{Yelp}} & \multicolumn{2}{c|}{\textbf{MNLI}} \\ \cline{3-8} 
 &  & \textbf{Pert\%} & \textbf{I\%} & \textbf{Pert\%} & \textbf{I\%} & \textbf{Pert\%} & \textbf{I\%} \\ \hline
\multirow{2}{*}{\textbf{BERT}} & Gen & \bf{6.5} & 1.04 & 11.6 & 1.5 & \bf{8.7} & \bf{1.9} \\
 & Ours & 6.7 &  \bf{1.02} & \bf{10.5} & \bf{1.49} & 9.2 & 2.1 \\ \hline
\multirow{2}{*}{\textbf{LSTM}} & Gen & 4.1 & 0.62 & 8.6 & 1.3 & 7.7 & 2.5 \\
 & Ours & \bf{3.19} & \bf{0.56} & \bf{6.2} & \bf{1.05} & \bf{8.2} & \bf{2.1} \\ \hline
\end{tabular}}}
 \caption{Comparison with Genetic Attack.}
 \label{table3b}
\end{subtable}
\begin{subtable}[h]{0.49\textwidth}

\centering
\resizebox{1.0\textwidth}{!}{%
{\renewcommand{\arraystretch}{1.2}
\begin{tabular}{|c|c|c|c|c|c|c|c|}
\hline
\multirow{2}{*}{\textbf{Model}} & \multirow{2}{*}{\textbf{Attack}} & \multicolumn{2}{c|}{\textbf{IMDB}} & \multicolumn{2}{c|}{\textbf{Yelp}} & \multicolumn{2}{c|}{\textbf{MNLI}} \\ \cline{3-8} 
 &  & \textbf{Pert\%} & \textbf{I\%} & \textbf{Pert\%} & \textbf{I\%} & \textbf{Pert\%} & \textbf{I\%} \\ \hline
\multirow{2}{*}{\textbf{BERT}} & PWWS & \bf{5.2} & \bf{0.74} & \bf{7.3} & \bf{1.5} & \bf{7.1} & 1.71 \\
 & Ours & 7.5 & 0.9 & 9.9 & 1.9 & 9.6 & \bf{1.48} \\ \hline
\multirow{2}{*}{\textbf{LSTM}} & PWWS & 2.3 & \bf{0.3} & \bf{4.8} & \bf{1.29} & \bf{6.6} & \bf{1.5} \\
 & Ours & \bf{1.9} & 0.4 & 5.5 & \bf{1.29} & 7.8 & 2.1 \\ \hline
\end{tabular}}}
 \caption{Comparison with PWWS.}
 \label{table3c}
\end{subtable}
\hfill
\begin{subtable}[h]{0.49\textwidth}
\centering
\resizebox{1.0\textwidth}{!}{%
{\renewcommand{\arraystretch}{1.2}
\begin{tabular}{|c|c|c|c|c|c|c|c|}
\hline
\multirow{2}{*}{\textbf{Model}} & \multirow{2}{*}{\textbf{Attack}} & \multicolumn{2}{c|}{\textbf{IMDB}} & \multicolumn{2}{c|}{\textbf{Yelp}} & \multicolumn{2}{c|}{\textbf{MNLI}} \\ \cline{3-8} 
 &  & \textbf{Pert\%} & \textbf{I\%} & \textbf{Pert\%} & \textbf{I\%} & \textbf{Pert\%} & \textbf{I\%} \\ \hline
\multirow{2}{*}{\textbf{BERT}} & TF & 9.0 & 1.21 & 5.2 & \bf{1.1} & 11.6 & \bf{1.23} \\
 & Ours & \bf{6.9} & \bf{0.9} & \bf{6.6} & 1.2 & \bf{11.4} &  1.41 \\ \hline
\multirow{2}{*}{\textbf{LSTM}} & TF & \bf{2.2} & 2.3 & 5.7 & {2.06} & \bf{9.8} & 1.7 \\
 & Ours & 2.4 & \bf{1.5} & \bf{5.3} & \bf{1.5} & 10.1 & \bf{1.4}\\ \hline
\end{tabular}}}
  \caption{Comparison with TextFooler (TF).}
  \label{table3d}
\end{subtable}
\caption{Result comparison. Pert\% is the perturbation and I\% is the average grammatical error increase.}
\label{table6}
\end{table*}
\subsection{Experimental Settings}
In a black box setting, the attacker has no access to the training data of the target model. Therefore, we made sure to train the attention models on a different dataset. For attacking the target model trained on IMDB, we trained our attention model on the Yelp Reviews and vice-versa. For entailment, we trained the attention model on SNLI~\citep{bowman2015large} and attacked the target model trained on MNLI. Following~\cite{jin2019bert}, the target models are attacked on same $1000$ samples, sampled from the test set of each dataset. The same set of samples are used across all baselines when evaluating on a single dataset. For entailment task we only perturb the premise and leave the hypothesis unchanged. We used spacy for POS tagging and filtered out stop words using NLTK. We used Universal Sentence Encoder~\cite{cer2018universal} to encode the perturbed inputs while performing LSH. The hyperparameters $d$ and $L$ are tuned on the validation set ($10\%$ of each dataset). Additional details regarding hyperparameter tuning and attention models can be found in appendix.
\subsection{Evaluation Metrics}
We use $(1)$ \emph{attack success rate} -- the ratio of the successful attacks to total number of attacks, $(2)$ \emph{query count} -- the number of queries, $(3)$ \emph{perturbation rate} -- the percentage of words substituted in an input and $(4)$ \emph{grammatical correctness} -- the average grammatical error increase rate (calculated using Language-Tool\footnote{https://languagetool.org/}) to verify the quality of generated adversarial examples. For all the metrics except attack success rate, lower the value better the result. Also, for all metrics,
we report the average score across all the generated adversarial examples on each dataset. Further, we also conducted human evaluation to assess the quality of generated adversarial examples.
\begin{figure*}
     \centering
     \begin{subfigure}[b]{0.45\textwidth}
         \centering
         \includegraphics[width=\textwidth, height=0.5\textwidth]{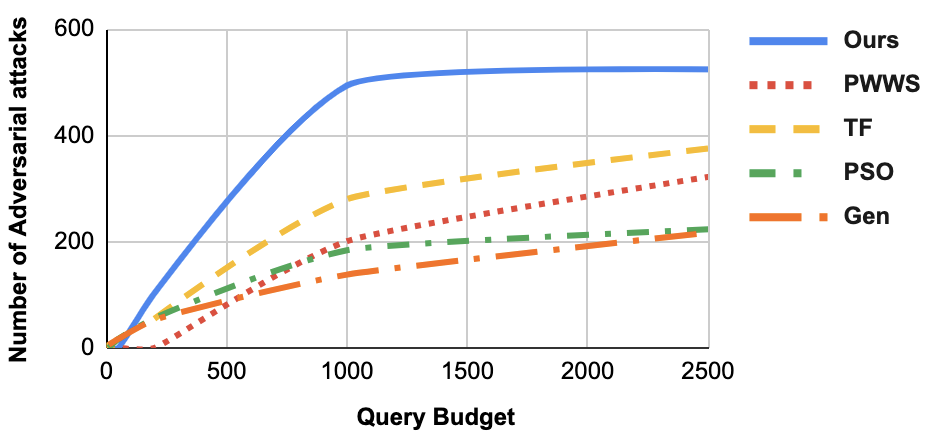}
         \caption{Adv. samples generated against BERT on IMDB}
         \label{fig:y equals x}
     \end{subfigure}
     \hfill
     \begin{subfigure}[b]{0.45\textwidth}
         \centering
         \includegraphics[width=\textwidth, height=0.5\textwidth]{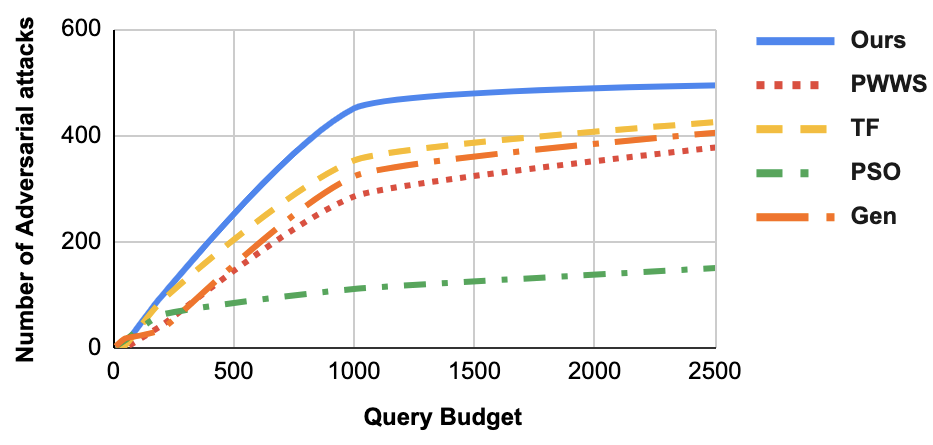}
         \caption{Adv. samples generated against BERT on Yelp}
         \label{fig:three sin x}
     \end{subfigure}
     \hfill
     \begin{subfigure}[b]{0.45\textwidth}
         \centering
         \includegraphics[width=\textwidth, height=0.5\textwidth]{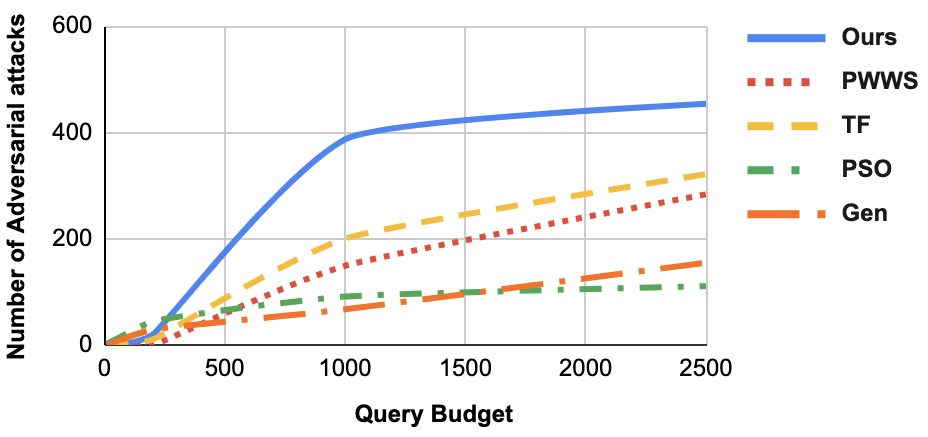}
         \caption{Adv. samples generated against LSTM on IMDB}
         \label{fig:five over x}
     \end{subfigure}
     \hfill
    \begin{subfigure}[b]{0.45\textwidth}
         \centering
         \includegraphics[width=\textwidth, height=0.5\textwidth]{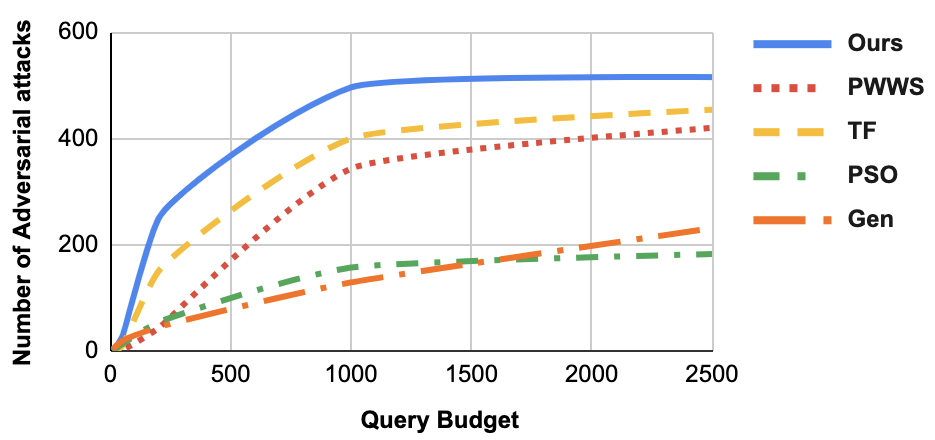}
         \caption{Adv. samples generated against LSTM on Yelp}
         \label{fig:five over x}
     \end{subfigure}
        \caption{Comparison of the number of adversarial samples generated by varying the query budget ${L}$.}
        \label{fig:lqs}
\end{figure*}
\begin{table*}
\centering
\resizebox{1.0\textwidth}{!}{%
{\renewcommand{\arraystretch}{1.0}
\begin{tabular}{|c|c|c|c|c|c|c|c|c|c|c|c|c|}
\hline
\multirow{2}{*}{\textbf{Dataset}} & \multicolumn{3}{c|}{\textbf{Random}} & \multicolumn{3}{c|}{\textbf{Only Attention}} & \multicolumn{3}{c|}{\textbf{Only LSH}} & \multicolumn{3}{c|}{\textbf{Both LSH and Attention}} \\ \cline{2-13} 
 & Suc\% & Pert\% & Qrs & Suc\% & Pert\% & Qrs & Suc\% & Pert\% & Qrs & Suc\% & Pert\% & Qrs \\ \hline
\textbf{IMDB} & 90.5 & 13.3 & 507.9 & 94.0 & 9.3 & 851.3 & 95.3 & 8.0 & 694.9 & \textbf{96.4} & \textbf{7.5} & \textbf{562.9} \\ \hline
\textbf{Yelp} & 87.3 & 15.0 & 305.9 & 91.0 & 11.0 & 550.0 & 90.2 & 10.2 & 475.2 & \textbf{92.6} & \textbf{9.8} & \textbf{366.2} \\ \hline
\textbf{MNLI} & 88.8 & 14.3 & 60.1 & 92.4 & 11.7 & 121.2 & 94.3 & 10.1 & 100.1 & \textbf{95.1} & \textbf{9.6} & \textbf{66.1} \\ \hline
\end{tabular}}}
\caption{Ablation Study of attention mechanism and LSH on PWWS search space.}
\label{table3}
\end{table*}
\subsection{Results}
Table \ref{table2} and \ref{table3} shows the comparison of our proposed method with each baseline across all evaluation metrics. On an average across all baselines, datasets and target models we are able to reduce the query count by $75\%$. The PSO and Genetic attack takes atleast $50$\emph{x} and $20$\emph{x} more queries respectively than our attack strategy. Also, when compared to PWWS and TF we are able to reduce the query count by atleast $65\%$ and $33\%$ respectively while compromising the success rate by less than $2.0\%$. The perturbation rate and the grammatical correctness is also within $1\%$ of the best baseline. In comparison to PSO and Genetic Attack, we are able to achieve even a lower perturbation rate and grammatical error rate with much lesser queries across some datasets and target models. Similarly, our attack outperforms TextFooler almost on all evaluation metrics. The runtime comparison and the anecdotes from generated adversarial text are provided in the appendix.
\section{Ablation Study}
We study the effectiveness of attention and LSH component in our method by doing a three way ablation. We observe the change in success rate, perturbation rate and queries when both or either one of the two ranking components are removed.

\textbf{No LSH and attention}: First, we remove both the attention and LSH scoring steps and rank the words in random order. Table \ref{table3} shows the results obtained on BERT across all three datasets. On an average the attack success rate drops by $7\%$, the perturbation rate increases drastically by $6\%$. This shows that although the query count reduces, substituting words in random order degrades the quality of generated adversarial examples and is not effective for attacking target models.

\textbf{Attention and no LSH}: We remove the LSH component of our ranking step and rank words based upon only the attention scores obtained from the attention model. Table \ref{table3} shows the results on BERT across all datasets. On an average the attack success rate drops by $2.5\%$, the perturbation rate increases by $3\%$ and the query increases by $37\%$. Therefore, LSH reduces the queries significantly by eliminating near duplicates in search space.

\textbf{LSH and no Attention}: We remove the attention component and rank words using only LSH. Results in Table \ref{table3} shows that on an average, without attention scoring the attack success rate drops by $2\%$, the perturbation rate increases by $0.5\%$ and the query increases by $20\%$. Therefore, attention is important as it not only reduces queries but also enables the ranking method to prioritize important words required in target model prediction.

\textbf{With LSH and Attention}: Finally in Table \ref{table3} we observe that, using both LSH and attention in our ranking our attack has a much better success rate, a lower perturbation rate in much lesser queries. This shows that both the components are necessary to do well across all evaluation metrics.\\
We obtained similar results on LSTM when evaluating across different datasets and search spaces.
\section{Quantitative Analysis}
     \begin{figure}[b]
         \centering
         \includegraphics[width=0.45\textwidth, height=0.25\textwidth]{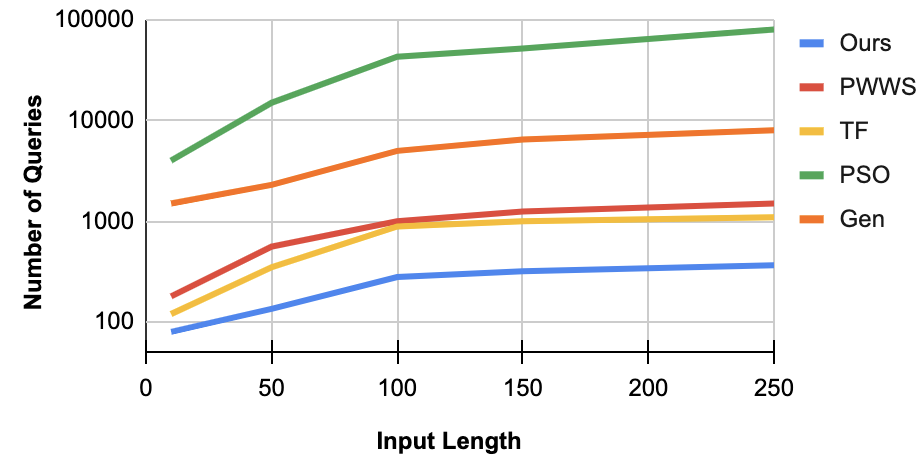}
         \caption{Queries taken vs number of words in input}
         \label{fig:qrs}
     \end{figure}

\subsection{Limited Query Setting}
In this setting, the attacker has a fixed query budget $L$, and can generate an attack in $L$ queries or less. To demonstrate the efficacy of our attack under this constraint, we vary the query budget $L$ and observe the attack success rate on BERT and LSTM across IMDB and Yelp datasets. We vary the query budget from $0$ to $2500$ and observe how many adversarial examples can be generated successfully on a test set of $500$ samples. We keep the search space same (used in PWWS) across all the search methods. The results in Figure \ref{fig:lqs} shows that with a query budget of $1000$, our approach generates atleast $200 \ (44.4\%)$ more adversarial samples against both BERT and LSTM on IMDB when compared to the best baseline. Similarly, on Yelp our method generates atleast $100 \ (25\%)$ more adversarial samples on BERT and LSTM when compared to the best baseline. This analysis shows that our attack has a much higher success rate in a limited query setting, thus making it extremely useful for real world applications.

\subsection{Input Length} To demonstrate that how our strategy scales with change in the input length (number of words in the input) compared to other baselines, we attacked BERT on Yelp. We selected inputs having number of words in the range of $10$ to $250$ and observed the number of queries taken by each attack method. Results in figure \ref{fig:qrs} shows that our attack takes the least number of queries across all input lengths. Further, our attack scales much better on longer inputs ($>$ $250$ words) as it is $2$\emph{x} faster than PWWS and TextFooler, $13$\emph{x} faster than Genetic attack and $133$\emph{x} faster than PSO.

\subsection{Transferability}
An adversarial example is said to be \emph{transferable}, if it is generated against one particular target model but is able to fool other target models as well. We implemented transferability on IMDB and MNLI datasets across two target models. The results are shown in Table \ref{table4}. Our transferred examples dropped the accuracy of other target models on an average by $16\%$.
\begin{table}[]
\centering
\resizebox{0.45\textwidth}{!}{%
{\renewcommand{\arraystretch}{1.0}
\begin{tabular}{|c|c|c|c|}
\hline
\textbf{Transfer} & \textbf{Accurracy} & \textbf{IMDB} & \textbf{MNLI} \\ \hline
\multirow{2}{*}{\textbf{BERT $\rightarrow$ LSTM}} & Original & 90.9 & 85.0 \\ \cline{2-4} 
 & Transferred & \textbf{72.9} & \textbf{60.6} \\ \hline
\multirow{2}{*}{\textbf{LSTM $\rightarrow$ BERT}} & Original & 88.0 & 70.1 \\ \cline{2-4} 
 & Transferred & \textbf{67.7} & \textbf{62.1} \\ \hline
\end{tabular}}}
\caption{Transferability on IMDB and MNLI datasets}
\label{table4}
\end{table}

\subsection{Adversarial Training}
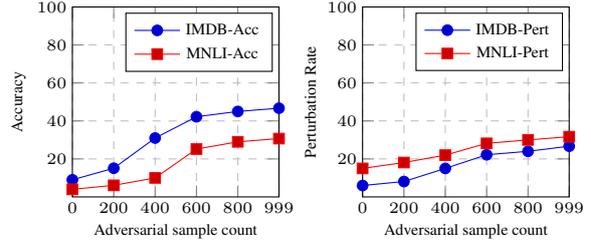
\begin{figure}
    \centering

\begin{tikzpicture}[trim left = -0.8cm]
\tiny
\usetikzlibrary{calc}
  \begin{axis}[name=plot1,    xlabel={Adversarial sample count},
    ylabel={Accuracy},
    xmin=0, xmax=999,
    ymin=0, ymax=100,
    xtick={0,200,400,600,800,999},
    ytick={20,40,60,80,100},
    legend pos=north east,
    ymajorgrids=true,
    xmajorgrids=true,
    grid style=dashed,
    height=4.1cm,width=4.3cm]
   \addplot
    coordinates {
    (0,9.0)(200,15.1)(400,31.0)(600,42.2)(800,45.0)(999,46.7)
    };
    \addlegendentry{IMDB-Acc}
\addplot
    coordinates {
    (0,4.0)(200,6.1)(400,10.0)(600,25.2)(800,29.0)(999,30.7)
    };
    \addlegendentry{MNLI-Acc}
  \end{axis}
  \begin{axis}[name=plot2,at={($(plot1.east)+(1.1cm,0)$)},anchor=west,
      xlabel={Adversarial sample count},
    ylabel={Perturbation Rate},
    xmin=0, xmax=999,
    ymin=0, ymax=100,
    xtick={0,200,400,600,800,999},
    ytick={20,40,60,80,100},
    legend pos=north east,
    ymajorgrids=true,
    xmajorgrids=true,
    grid style=dashed,
    height=4.1cm,width=4.3cm]
    \addplot
    coordinates {
    (0,6.0)(200,8.1)(400,15.0)(600,22.2)(800,24.0)(999,26.7)
    };
    \addlegendentry{IMDB-Pert}
\addplot
    coordinates {
    (0,15.0)(200,18.1)(400,22.0)(600,28.2)(800,30.0)(999,31.7)
    };
    \addlegendentry{MNLI-Pert}
  \end{axis}
\end{tikzpicture}
\caption{Increase in after attack accuracy and perturbation rate as more adversarial samples are augmented.}
\label{fig:adlearn}
\end{figure}

We randomly sample $10\%$ samples from the training dataset of MNLI and IMDB and generate adversarial examples using our proposed strategy.
We augmented the training data with the generated adversarial examples
and re-trained BERT on IMDB and MNLI tasks. We then again attacked BERT with our proposed strategy and observed the changes. The results in Figure \ref{fig:adlearn} shows that as we add more adversarial examples to the training set, the model becomes more robust to attacks. The after attack accuracy and perturbation rate increased by $35\%$ and $17\%$ respectively and required higher queries to attack.
\section{Qualitative Analysis}
\subsection{Human Evaluation} We verified the quality of generated adversarial samples via human based evaluation as well. We asked the evaluators to classify the adversarial examples and  score them in terms of grammatical correctness (score out of $5$) as well as its semantic similarity compared to the original text. We randomly sampled $25\%$ of original instances and their corresponding adversarial examples generated on BERT for IMDB and MNLI datasets on PWWS search space. The actual class labels of adversarial examples were kept hidden and the human judges were asked to classify them. We also asked the human judges to evaluate each sample for its semantic similarity and assign a score of $0$, $0.5$ or $1$ based on how well the adversarial examples were able to retain the meaning of their original counterparts. We also asked them to score each example in the range $1$ to $5$ for grammatical correctness. Each adversarial example was evaluated by $3$ human evaluators and the scores obtained were averaged out. The outcome is in Table \ref{table5}.
\begin{table}[h!]
\centering
\small
{\renewcommand{\arraystretch}{1.2}
\begin{tabular}{|c|c|c|}
\hline
\textbf{Evaluation criteria} & \textbf{IMDB} & \textbf{MNLI} \\ \hline
{Classification result} & 94\% & 91\% \\ \hline
{Grammatical Correctness} & 4.32 & 4.12 \\ \hline
{Semantic Similarity} & 0.92 & 0.88 \\ \hline
\end{tabular}}
\caption{Demonstrates scores given by evaluators}
\label{table5}
\end{table}
\section{Conclusion}
We proposed a query efficient attack that generates plausible adversarial examples on text classification and entailment tasks. Extensive experiments across \emph{three} search spaces and \emph{four} baselines shows that our attack generates high quality adversarial examples with significantly lesser queries. Further, we demonstrated that our attack has a much higher success rate in a limited query setting, thus making it extremely useful for real world applications.
\section{Future Work}
Our proposed attack provides a strong baseline for more query efficient black box attacks. The existing word level scoring methods can be extended to sentence level. Also, the attention scoring model used can be trained on different datasets to observe how the success rate and the query efficiency gets affected. Furthermore, existing attack methods can be evaluated against various defense methods to compare the effectiveness of different attacks.
\section{Acknowledgement}
We would like to thank all the reviewers for their critical insights and positive feedback. We would also like to thank Riyaz Ahmed Bhat for having valuable discussions which strengthen our paper and helped us in responding to reviewers.
\bibliography{anthology,custom,references}
\bibliographystyle{acl_natbib}
\appendix
\clearpage
\newpage
\section{Appendix}
\label{sec:appendix}
\begin{algorithm}[h!]
\caption{Word Ranking}
\algorithmicrequire{ Test sample $\mathcal{X}$}\\
\algorithmicensure{ $\mathcal{W}$ containing score of each word $x_i$}
\begin{algorithmic}[1]
\State $F_{attn}\ \gets HAN() \ or \ DA()$\;
\State $\alpha \gets  F_{attn}(\mathcal{X})$\;
\For{$x_i \ \textbf{in} \ \mathcal{X}$}
\State ${S} \gets\ Synonyms(x_{i})$
\For{$w_j \ \textbf{in} \ S$}
\State $\mathcal{X}_{ij} \gets Replace \ x_i \ with \ w_j \ in \ \mathcal{X}$
\State $\mathcal{V}_j \gets encode(\mathcal{X}_{ij})$
\State $\{b_1..b_\mathcal{K}\}=  \mathbf{LSH}(\mathcal{V}_j,\mathcal{X}_{ij})$
\EndFor
\For{$k = 1 \ to \ \mathcal{K}$}
\State $\mathcal{V}_k^*, \mathcal{X}_k^* \gets sample(b_k)$
\State $\Delta P_k = \mathbf{F}(\mathcal{X}) - \mathbf{F}(\mathcal{X}_k^*)$ 
\EndFor
\State $P_i = \max (\Delta P_k)$
\State $score_i \gets \alpha_i * P_i$
\State $\mathcal{W}.insert((score_i,x_i))$
\EndFor
\State  $Sort\ \mathcal{W} \ by \ score_i \ in \ descending \ order$\ \;
\end{algorithmic}
\end{algorithm}
     \begin{figure}[b]
         \centering
         \includegraphics[width=0.45\textwidth, height=0.25\textwidth]{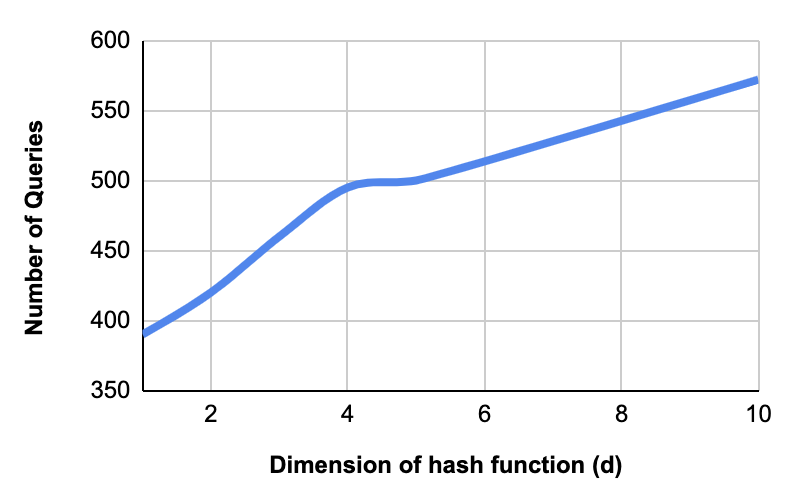}
         \caption{Queries vs the dimension $d$ of hash function}
         \label{d_queries}
     \end{figure}
     \begin{figure}[b]
         \centering
         \includegraphics[width=0.45\textwidth, height=0.25\textwidth]{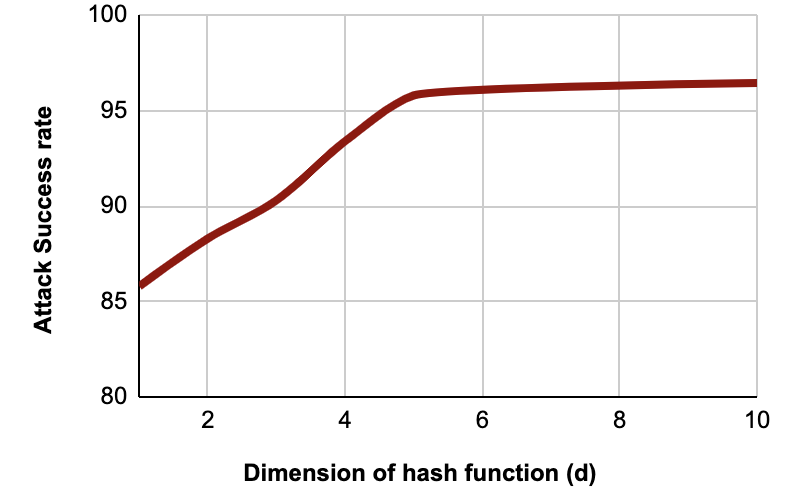}
         \caption{Success rate vs dimension $d$ of hash function}
         \label{d_success_rate}
     \end{figure}
\begin{table}[h!]
\centering
\small
{\renewcommand{\arraystretch}{0.9}
  \begin{tabular}{|l|c c c c|}
  \hline
    \textbf{Dataset} & \textbf{Train} 
      & \textbf{Test}& \textbf{Classes}& \textbf{Avg. Len} 
      \\
      \hline
     IMDB & 12K & 12K & 2 & 215 \\
     Yelp & 560K & 18K  &2 & 152\\
     MultiNLI  & 12K & 4K & 3 & 10\\
     \hline
  \end{tabular}}
  \caption{Statistics of all datasets}
  \label{table:6}
\end{table}

\subsection{Models}
\textbf{Target Models:} We attacked WordLSTM and BERT-base-uncased for evaluating our attack strategy on text classification and entailment tasks. For WordLSTM a single layer bi-directional LSTM with $150$ hidden units, $200$ dimensional GloVe vectors and a dropout of $0.3$ were used.\\
\textbf{Ranking Models:} We used Hierarchical Attention Networks (HAN) and Decompose Attention Model (DA) for classification and entailment tasks respectively. For training HA, we used 200 dimensional word2vec embeddings and 50 dimensinal GRU cells. We used $100$ dimensional word context vectors initialized  randomly. We trained the model with a batch size of $64$, learning rate of $0.0001$, dropout of $0.3$ and momentum of $0.9$. For DA, a $2$ layer LSTM with $200$ neurons and $200$ dimensional glove embeddings are used. A batch size of $32$, learning rate of $0.025$, dropout of $0.2$ are used. All the hyper-parameters were tuned on the $20\%$ validation set of each dataset.
\begin{table}[h!]
\centering
\small
\begin{tabular}{|c|c|}
\hline
\textbf{Attack} & \textbf{Runtime} \\ \hline
PSO & 72 hrs \\ \hline
Genetic Attck & 10 hrs \\ \hline
PWWS & 3 hrs \\ \hline
TextFooler & 2.5 hrs \\ \hline
Ours & 1 hrs \\ \hline
\end{tabular}
\caption{Runtime comparison while attacking BERT trained on Yelp dataset on a set of $500$ samples across WordNet search space.}
\end{table}
\subsection{Hyperparameter Study}
\begin{figure*}
     \begin{subfigure}[b]{0.45\textwidth}
         \centering
         \includegraphics[width=\textwidth, height=0.55\textwidth]{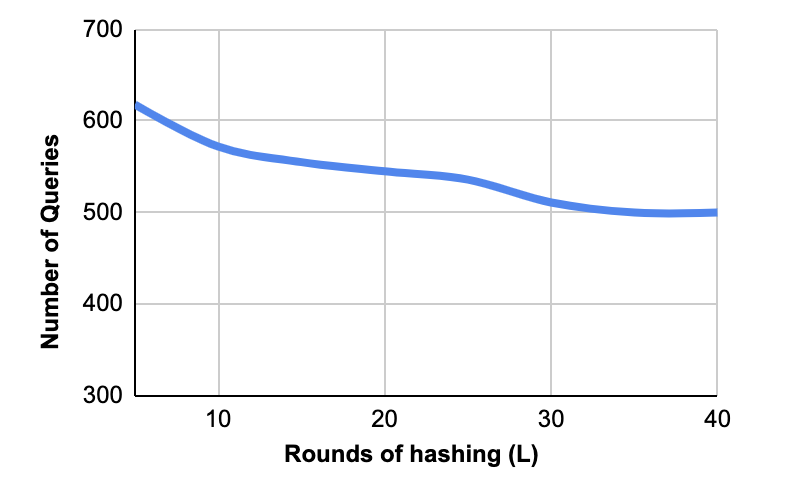}
         \caption{Queries vs rounds of hashing $L$}
         \label{l_queries}
     \end{subfigure}
     \hfill
    \begin{subfigure}[b]{0.45\textwidth}
         \centering
         \includegraphics[width=\textwidth, height=0.55\textwidth]{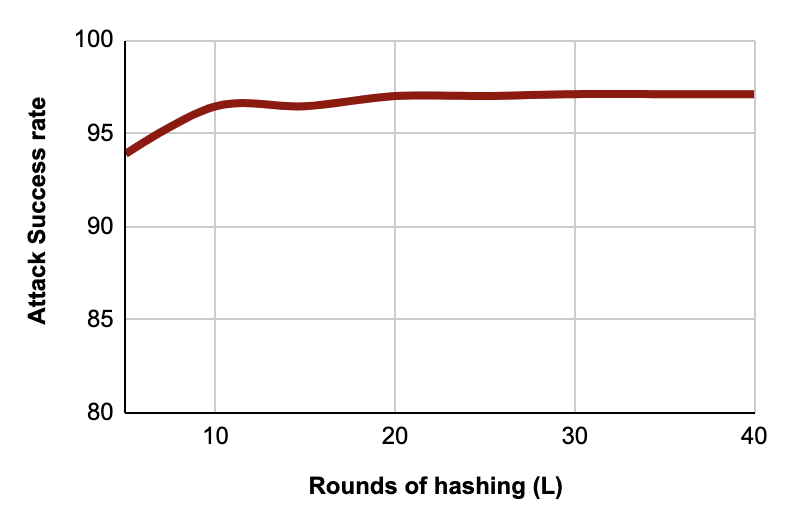}
         \caption{Success rate vs rounds of hashing $L$}
         \label{l_success_rate}
     \end{subfigure}
        \caption{Comparison of success rate and queries required by changing the rounds of hashing $L$.}
        \label{fig:lqs}
\end{figure*}

Figure \ref{d_queries} and \ref{d_success_rate} shows the variation of attack success rate and queries taken to attack the target model as $d$ increases. With increase in $d$ the number of collisions decreases and therefore the number of buckets obtained $\mathcal{K}$ increases. This increases the overall query count.
Also, with increase in $d$ the success rate first increases and then remains unchanged. Therefore, we use $d = 5$ because after that the success rate is almost the same but the query count increases drastically.
Figure \ref{l_queries} and \ref{l_success_rate} shows the variation of attack success rate and queries taken to attack the target model as the rounds of hashing $L$ increases. Conducting multiple rounds of hashing reduces the probability that that similar perturbed text inputs are mapped to different buckets. We choose $L = 15$ as after it the attack success rate and the queries remain almost unchanged. The values of $d$, $L$ are kept same across all datasets and target models.

\definecolor{ggreen}{HTML}{3F9C3E}
\begin{table*}[h!]
\centering
\small
{\renewcommand{\arraystretch}{1.15}
\begin{tabular}{|p{11cm}|p{3.2cm}|}
\hline
\textbf{Examples} & \textbf{Prediction} \\ \hline
The movie has an excellent screenplay (the situation is credible, the action has pace), \textbf{\textcolor{ggreen} {first-class}} \textbf{[\textcolor{red}{fantabulous}]} direction and acting (especially the 3 leading actors but the others as well -including the mobster, who does not seem to be a professional actor). I \textbf{\textcolor{ggreen}{wish}} \textbf{[\textcolor{red}{want}]} the movie, the director and the actors success. & \textbf{{\textcolor{ggreen}{Positive}} $\xrightarrow[]{}$ \textbf{\textcolor{red}{Negative}}} \\
\hline
	Let me start by saying I don't recall laughing once during this comedy. From the opening scene, our protagonist Solo (Giovanni Ribisi) shows himself to be a self-absorbed, feeble, and neurotic loser completely unable to cope with the smallest responsibilities such as balancing a checkbook, keeping his word, or forming a coherent thought. I guess we're supposed to be drawn to his fragile vulnerability and cheer him on through the process of clawing his way out of a deep depression. I actually \textbf{\textcolor{ggreen}{wanted}} \textbf{[\textcolor{red}{treasured}]} him to get his kneecaps busted at one point. The dog was not a character in the film. It was simply a prop to be used, neglected, scorned, abused, coveted and disposed of on a whim. So be warned. & \textbf{{\textcolor{ggreen}{Negative}} $\xrightarrow[]{}$ \textbf{\textcolor{red}{Postive}}} \\
\hline
Local-international \textbf{\textcolor{ggreen}{gathering}} \textbf{[\textcolor{red}{assembly}]} \textbf{\textcolor{ggreen}{spot}} \textbf{[\textcolor{red}{stain}]} since the 1940s. One of the coolest pubs on the planet. Make new friends from all over the world, with some of the \textbf{\textcolor{ggreen}{best}} {\textbf{[\textcolor{red}{skilful}]}} regional and imported beer selections in town. & \textbf{{\textcolor{ggreen}{Postive}} $\xrightarrow[]{}$  \textbf{\textcolor{red}{Negative}}} \\
\hline
This film is strange, even for a silent movie. Essentially, it follows the adventures about a engineer in post-revolutionary Russia who daydreams about going to Mars. In this movie, it seems like the producers KNOW the Communists have truly screwed up the country, but also seems to want to make it look like they've accomplished something good. Then we get to the "Martian" scenes, where everyone on Mars wears goofy hats. They have a revolution after being inspired by the Earth Men, but are quickly betrayed by the Queen who sides with them. Except it's all a dream, or is it. (And given that the Russian Revolution eventually lead to the Stalin dictatorship, it makes you wonder if it was all allegory.) \textbf{\textcolor{ggreen}{Now}} \textbf{[\textcolor{red}{Nowdays}]}, I've seen GOOD Russian cinema. For instance, Eisenstein's Battleship Potemkin is a good movie. This is just, well, silly.
& \textbf{{\textcolor{ggreen}{Negative}} $\xrightarrow[]{}$ \textbf{\textcolor{red}{Postive}}} \\
\hline
This movie is one of the \textbf{\textcolor{ggreen}{worst}} \textbf{[\textcolor{red}{tough}]} \textbf{\textcolor{ggreen}{remake}} \textbf{[\textcolor{red}{remakes}]} I have ever seen in my \textbf{\textcolor{ggreen}{life}} \textbf{[\textcolor{red}{aliveness}]}! The acting is \textbf{\textcolor{ggreen}{laughable}} \textbf{[\textcolor{red}{funnny}]} and Corman has not improved his piranhas any since 1978. 90\% of the \textbf{\textcolor{ggreen}{special}} \textbf{[\textcolor{red}{exceptional}]} \textbf{\textcolor{ggreen}{effects}} \textbf{[\textcolor{red}{impressions}]} are \textbf{\textcolor{ggreen}{taken}} \textbf{[\textcolor{red}{lifted}]} from Piranha (1978), Up From The Depths (1979) and Humanoids From The Deep (1979). It makes Piranha II: The Spawning look like it belongs on the American Film Institute List. & 
\textbf{{\textcolor{ggreen}{Negative}} $\xrightarrow[]{}$ \textbf{\textcolor{red}{Postive}}} \\
\hline
 The story \textbf{\textcolor{ggreen}{ultimately}} \textbf{[\textcolor{red}{eventually}]} takes hold and grips hard. & \textbf{{\textcolor{ggreen}{Positive}} $\xrightarrow[]{}$ \textbf{\textcolor{red}{Negative}}} \\ 
\hline
It's weird, wonderful, and not \textbf{\textcolor{ggreen}{neccessarily}} \textbf{[\textcolor{red}{definitely}]} for kids. & {{ \textbf{\textcolor{ggreen}{Negative}} $\xrightarrow[]{}$ \textbf{\textcolor{red}{Positive}}}}. \\ 
\hline
\end{tabular}}
  \caption{Demonstrates adversarial examples generated after attacking BERT on classification task. The actual word is highlighted green and substituted word is in square brackets colored red. Prediction shows before and after labels marked green and red respectively.}
  \label{table:6}
\end{table*}

\definecolor{ggreen}{HTML}{3F9C3E}
\begin{table*}[]
\small
{\renewcommand{\arraystretch}{1.15}
\begin{tabular}{|p{11cm}|p{3.2cm}|} \hline
\textbf{Examples} & \textbf{Prediction} \\ \hline
\textbf{Premise}:  If we travel for 90 minutes, we could \textbf{\textcolor{ggreen}{arrive}} \textbf{[\textcolor{red}{reach}]} arrive at larger ski resorts.  & \multirow{2}{*}{{ \textbf{\textcolor{ggreen}{Entailment}} $\xrightarrow[]{}$ \textbf{\textcolor{red}{Neutral}}}} \\
\textbf{Hypothesis}: Larger ski resorts are 90 minutes away. & \\
\hline
\textbf{Premise}: I should put it in this \textbf{\textcolor{ggreen}{way}} \textbf{[\textcolor{red}{manner}]}. & \multirow{2}{*}{{ \textbf{\textcolor{ggreen}{Entailment}} $\xrightarrow[]{}$ \textbf{\textcolor{red}{Neutral}}}} \\
\textbf{Hypothesis}: I'm not explaining it. & \\
\hline
\textbf{Premise}: \textbf{\textcolor{ggreen}{Basically}} \textbf{[\textcolor{red}{Crucially}]}, to sell myself. & \multirow{2}{*}{{ \textbf{\textcolor{ggreen}{Contradict}} $\xrightarrow[]{}$ \textbf{\textcolor{red}{Neutral}}}}  \\
\textbf{Hypothesis}: Selling myself is a very important thing.  &\\
\hline
\textbf{Premise}: \textbf{\textcolor{ggreen}{June}} \textbf{[\textcolor{red}{April}]} 21, 1995, provides the specific requirements for assessing and reporting on controls. & \multirow{2}{*}{{ \textbf{\textcolor{ggreen}{Contradict}} $\xrightarrow[]{}$ \textbf{\textcolor{red}{Entail}}}} \\
\textbf{Hypothesis}: There are specific requirements for assessment of legal services. & \\
\hline
\end{tabular}}
  \caption{Demonstrates adversarial examples generated after attacking BERT on entailment task. The actual word is highlighted green and substituted word is in square brackets colored red. Prediction shows before and after labels marked green and red respectively.}
  \label{table:7}
\end{table*}

\end{document}